\documentclass[letterpaper, 10 pt, conference]{ieeeconf}  

\IEEEoverridecommandlockouts                              
\usepackage[utf8]{inputenc} 
\usepackage[T1]{fontenc}    
\usepackage{hyperref}       
\usepackage{url}            
\usepackage{booktabs}       
\usepackage{amsfonts}       
\usepackage{nicefrac}       
\usepackage{microtype}      
\usepackage{caption}
\usepackage{subcaption}
\usepackage{graphicx}
\usepackage{float}
\usepackage{color}
\usepackage{amsmath}
\usepackage[normalem]{ulem}
\useunder{\uline}{\ul}{}

\title{\LARGE \bf
Learning to Sequence Robot Behaviors for Visual Navigation
}

\author{Hadi Salman, Puneet Singhal, Tanmay Shankar, Peng Yin, Ali Salman, William Paivine,  \\Guillaume Sartoretti, Matthew Travers, and Howie Choset
\thanks{H. Salman, P. Singhal, T. Shankar, W. Paivine, M. Travers, G. Sartoretti, and H. Choset are with The Robotics Institute at Carnegie Mellon University, Pittsburgh, PA 15213, USA
        {\tt\small \{hadis, psinghal, tshankar, wjp, mtravers\}@andrew.cmu.edu, \{gsartore,choset\}@cs.cmu.edu}.}%
\thanks{P. Yin is with the Chinese Academy of Sciences, Beijing
        {\tt\small yinpeng@sia.cn}.}%
\thanks{A. Salman is with the Lebanese University, Hadat, Lebanon
        {\tt\small alicsalman1@st.ul.edu.lb}.}%
}

\begin{document}

\maketitle
\thispagestyle{empty}
\pagestyle{empty}

\begin{abstract}
Recent literature in the robotics community has focused on learning robot behaviors that abstract out lower-level details of robot control. 
To fully leverage the efficacy of such behaviors, it is necessary to select and sequence them to achieve a given task. 
In this paper, we present an approach to both learn and sequence robot behaviors, applied to the problem of visual navigation of mobile robots. 
We construct a layered representation of control policies composed of low-level behaviors and a meta-level policy. The low-level behaviors enable the robot to locomote in a particular environment while avoiding obstacles, and the meta-level policy actively selects the low-level behavior most appropriate for the current situation based purely on visual feedback. 
We demonstrate the effectiveness of our method on three simulated robot navigation tasks: a legged hexapod robot which must successfully traverse varying terrain, a wheeled robot which must navigate a maze-like course while avoiding obstacles, and finally a wheeled robot navigating in the presence of dynamic obstacles. 
We show that by learning control policies in a layered manner, we gain the ability to successfully traverse new compound environments composed of distinct sub-environments, and outperform both the low-level behaviors in their respective sub-environments, as well as a hand-crafted selection of low-level policies on these compound environments. 
\end{abstract}

\section{Introduction}
During autonomous deployments, mobile robots require the capability to react online to changes in their surrounding, such as terrain changes, dynamic obstacles, etc.
A common way to approach this challenge is to define high-level robot behaviors, and to endow robots with the possibility to switch between such behaviors based on their environment~\cite{Nicolescu:2002:HAB:544741.544798, richter2012robotic, 10.3389/fncom.2014.00062}.
In this paper, we consider the task of improving the navigation of different mobile robots, by sequencing learned or designed behaviors based on visual feedback.
Traditional planning methods to solve this problem rely on hand-crafted state representations and heuristics for planning, which often fail to generalize to new scenarios~\cite{richter2012robotic}.
Inspired by recent results in machine learning, where deep neural networks can model complex control policies directly from raw inputs~\cite{mnih2015human, mnih2016asynchronous, silver2016mastering, koutnik2014evolving}, we propose to cast this problem in a reinforcement learning (RL) framework. 
Our main contribution is a hierarchical RL framework, where agents are trained to both learn and sequence robot behaviors for autonomous navigation by relying solely on raw visual input from a monocular camera.
In this framework, agents learn low-level locomotive behaviors, while meta-agents explore the use of these behaviors in different scenarios to maximize the distance travelled through the environment while avoiding obstacles.
\begin{figure}[!t]
	\includegraphics[width=0.5\textwidth]{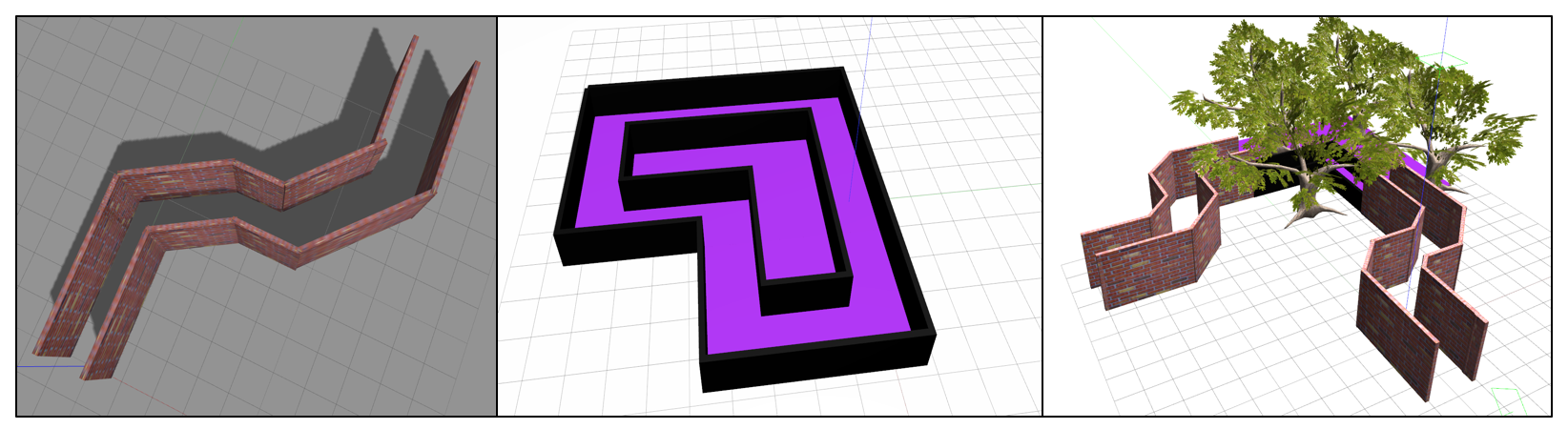}
	\caption{Depiction of elementary environments~1 (left) and environment~2 (center), as well as a composed environment (right) constructed from combining these environments.}
	\label{fig:envs_turtlebot}
\end{figure}
Specifically, we rely on deep Q-learning \cite{mnih2015human} to teach the robot low-level behaviors, as well as to train a meta-agent to sequence the robot's behaviors. Each of the low-level behaviors allow the robot to locomote while avoiding obstacles  in a given environment based on raw visual feedback. These low-level behaviors can be complex gaits, or can be learnt policies that execute simple actions such as forward motion or left and right turns. 
We also maintain a \textit{meta-level policy} that selects the most appropriate low-level behavior for the current situation.




We present results of our hierarchical approach on both wheeled and legged robots in simulation. Our low-level behaviors are tailored to 
a specific environment, each with uniform appearance or structure (such as textured walls, rough terrain, etc.). In contrast, the meta-level policies are learnt in environments composed of several of the training appearances and terrains as shown in Fig.~\ref{fig:envs_turtlebot}. 
We show how the robots are able to navigate these novel environments by sequencing the appropriate lower-level behaviors based on their immediate surroundings. 
We further show how learning to sequence low-level behaviors results in a more effective overall policy than either of the individual sub-policies, even in the respective environments they were designed/trained for. 

This paper is organized as follows: Section \ref{related_work} provides related work to this paper. 
In Section \ref{approach}, we present our framework that uses hierarchical reinforcement learning for sequencing behaviors given only visual input in a navigation task.
In Section \ref{results}, we present the results of our framework and compare it with simple deep Q-network (DQN) architecture. In Section \ref{conclusion}, we conclude by summarizing our results and stating our future work.

\begin{figure*}[t]
	\includegraphics[width=\textwidth]{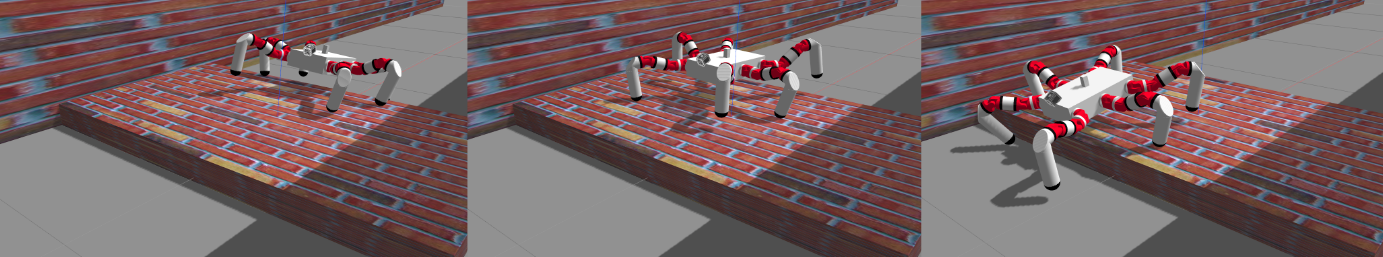}
	\caption{Depiction of the 6-legged robot traversing a tall obstacle by sequencing two gaits based on raw camera pixels.}
	\label{fig:snakemonster}
\end{figure*}

\section{Related Work}
\label{related_work}

Several efforts in the robotic community have focused on learning primitive-actions or skills; we provide a brief review of such works, and some recent developments that have taken steps towards sequencing these skills.

Skill Trees \cite{Konidaris11d}, introduced method to segment demonstrations into skills and \textit{chain} these skills. 
The problem of hierarchical reinforcement learning has been of interest for a considerable amount of time \cite{Sutton99betweenmdps}. 
In hierarchical RL, multi-step action policies are represented as \textit{options}; a meta-controller then selects which \textit{option} to apply. 
\cite{kulkarni2016hierarchical} developed an approach to hierarchical RL that provides ``intrinsic motivation'' to agents to perform certain subtasks, drastically helping the agents ability to achieve overall task completion. 

\cite{10.3389/fncom.2014.00062} constructed a layered approach to adapt, select, and sequence DMPs. \cite{DBLP:journals/corr/AndreasKL16} provided annotations of task structure, and optimized for overall task completion over a set of modular subpolicies. 
\cite{tlpkICRA17} built models of the preconditions and effects of parameterized skills. 
While \cite{xie2017towards} address monocular vision based navigation via reinforcement learning, they do not make use of existing robot behaviors. 
We note that while our paradigm of learning a \textit{meta-level policy} to sequence behaviors is also adopted by \cite{2017arXiv171009767F}, they do not address the challenge of learning from visual inputs.
\section{Approach}
\label{approach}
The problem of sequencing a set of robot behaviors from visual feedback can be posed as a Markov Decision Process (MDP) with \textit{temporal abstractions} inspired by \cite{Sutton99betweenmdps}. At a given state $s_t \in S$ corresponding to time step $t$, the robot chooses a behavior (i.e. a low level policy) $\pi_i$ from a predetermined set of behaviours $\Pi=\{\pi_1\dots \pi_n\}$ according to a meta policy $\Omega(s_t)$, and follows this behaviour $\pi_i$ for $N$ time steps. During these $N$ steps, an action $a_t \in \mathcal{A}$ is chosen according to the low level policy $\pi_i$ resulting in a new state $s_{t+1}$ of the robot and a collected reward $r_t \in \mathbb{R}$. The goal is to learn the meta-level policy $\Omega$ that sequentially chooses a low level policy $\pi_i$ every $N$ steps to maximize the cumulative reward~$ \mathcal{R}= \big[ \sum_{\tau=t}^{\infty}\gamma ^{\tau - t}r_{\tau} \big]$. 

In order to learn the meta-level policy $\Omega$ (and the low-level policies $\pi_i$ for $i \in {1 \dots n}$ in certain cases), we make use of Deep Q-learning \cite{mnih2015human}. Q-learning estimates the Q-values of state-action pair $(s,a)$, which is defined as the expected cumulative reward upon taking action $a$ from state $s$, and following policy $\pi$ thereafter. Formally, the Q-value of a state- action pair can be written as,
\begin{align}
Q(s,a) &= \mathbb{E}_{\pi} \big[ \mathcal{R} \big | s_t = s, a_t = a] 
\end{align}
In particular, Q-learning is a temporal difference method that optimizes the following loss function defined as, 
\begin{equation}\label{eq:lossFunction}
L_t = (r_t + \gamma \max_{a'} Q(s_{t+1},a') - Q(s_t,a_t))^2
\end{equation}
Deep Q-learning employs deep neural networks as function approximators to estimate these Q-values. For more details, we refer the reader to \cite{mnih2015human}.

\begin{figure*}[t]
	\includegraphics[width=\textwidth]{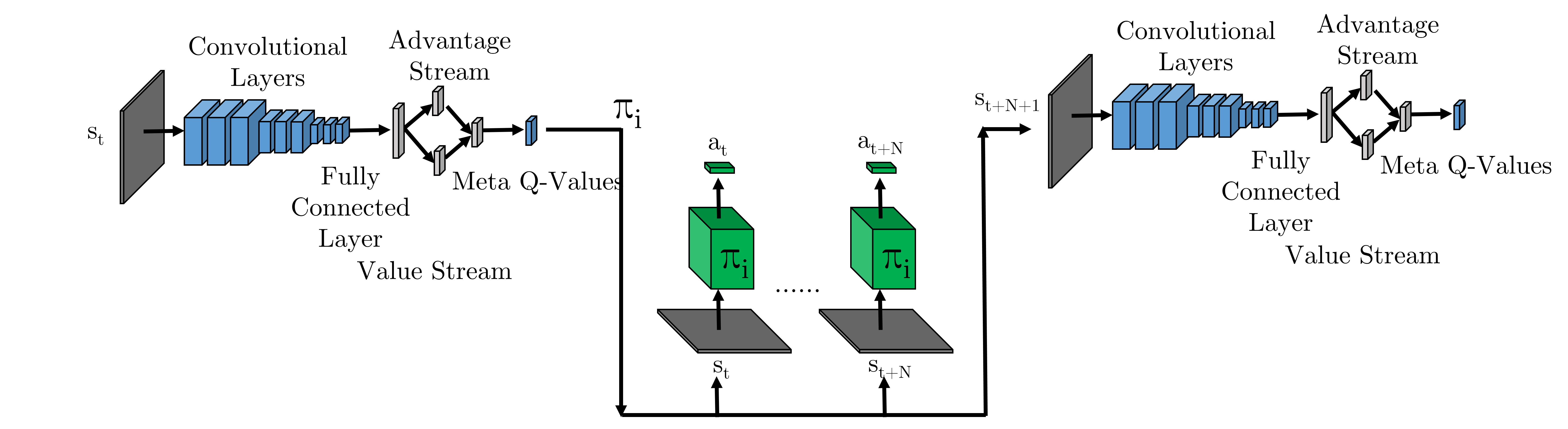}
	\caption{Schematic architecture of our approach. Here, we jointly depict the architecture of the Meta policy (Duelling DQN), as well as the low-level policies $\pi_i$. We further depict the schematic of calling the meta policy first, choosing a low-level control policy to execute, running the low-level policy for $N$ steps, and then subsequently using the meta level policy again to choose another low level policy. }	
	\label{fig:arch}
\end{figure*}

\subsection{Framework}
We consider two robot agents in this paper, each equipped with a monocular camera that serves to provide an observation of state $s_t$ of the robot. 
The first robot, a six-legged robot, must navigate an environment that consists of varying \textit{terrain}. 
The second robot, a differetial-drive robot, Turtlebot, must navigate an environment consisting of \textit{visually} dissimilar regions. 
In both cases, the objective of the robot is to maximize the distance it travels in the environment, while reconciling with changes in the terrain or the appearance of the environment, and avoiding obstacles (in the case of the Turtlebot).
Our reward function encodes this objective by positively rewarding the distance travelled by the robot and negatively penalizing any collisions with obstacles. 


Key to our approach is the idea of hierarchical reinforcement learning, where an agent learns control policies in a hierarchical framework. In our approach we consider two such levels:

\begin{enumerate}
    \item The \textit{low-level control policies} select robot actions (such as moving forward or turning left or right) on the basis of perceptual input (i.e. raw camera input). 

    \item At the higher level, there is a \textit{meta-level policy}, which selects which of these lower level policies to apply over an extended period of time. 
\end{enumerate}

The use of such a hierarchy of policies is augmented by constructing \textit{compound} environments that are combinations of several 
dissimilar \textit{elementary} environments depicted in Figure~\ref{fig:envs_turtlebot}. 
One low-level policy is trained to navigate each elementary environment, while the meta-level policy is trained to sequence these low-level policies to navigate the \textit{compound} environment. 
We note that the low-level policies only observe \textit{one} of the elementary environments during training (or are only designed to navigate one type of terrain in the case of the legged robot), and hence perform poorly on the other elementary environments. 
It is thus necessary for the meta-level policy to alternate between these low-level policies employed in order to successfully navigate the compound environment. 


We present a schematic of this hierarchical framework in Figure~\ref{fig:arch}.  The meta-level policy (depicted in blue) selects a control policy (in green), which then provides low-level commands to the robot.  
This combination of a hierarchical representation of policies combined with the notion of environments composed of distinct elementary components necessitates the use of a hierarchical framework. 
We provide a description of the low-level policies specific to each of these robots below, followed by a description of the meta-level policy. 


\subsection{Lower-level control policies}\label{section:lowlevelpolicies}
The low-level control policies enable the robot to locomote in their respective environments. 
We describe the form of the low-level policies for each of the robots we use.

\noindent

\subsubsection{Turtlebot}
We provide the Turtlebot with actions to move forward, and turning towards the left or right. 
The low-level policy must specify which of these actions the Turtlebot must take.
To evaluate the quality of each of these actions, we make use of a DQN, as described in Section \ref{approach}, to map an observation of state of the robot, $s_t$ to an estimate of the $Q$ values of these actions $Q(s,a, \theta) \ \forall a \in A$.  
In this paper, we use a variant of the DQN, i.e. a double duelling deep Q-network \cite{DBLP:journals/corr/WangFL15} for this. 
The low-level policy $\pi$ is then derived by acting greedily with respect to the estimate of $Q$ values, i.e., $\pi(s) = \arg\max_a Q(s,a,\theta)$. 
We train two instances $\pi_1$ and $\pi_2$ of the low-level control policy  in visually differing elementary environments.

\subsubsection{Legged Robot}
 In case of the 6-legged robot navigation, our low-level policies take the form of two different gaits. The first gait takes low steps at a relatively fast pace, ideal for traversing flat terrain quickly. The second gait takes higher steps at a slower pace, which is suitable for traversing obstacles present in the environment. We make use of Central Pattern Generators (CPG) \cite{sartoretticentral} to generate these gaits; overloading notation and referring to these gaits as $\pi_1, \pi_2$. To quickly traverse a compound environment, the legged robot would ideally use an appropriate combination of these gaits. 



We note that the actions provided to each of the robots are executed by underlying controllers that incorporate the dynamics of the robot - for example, the \textit{left} command steers the turtlebot forward by rotating its wheels at varying speeds. Also we note that we assume we do not have access to the dynamics of the robots, rather we use a physics simulator. Specifically, we use gym-gazebo \cite{zamora2016extending}, which is an extension of the OpenAI Gym \cite{1606.01540} for robotics that is based on the Robot Operating System (ROS) \cite{Quigley09} and the Gazebo simulator \cite{koenig2006gazebo}.


\begin{figure*}[t]
	\centering
	\includegraphics[width=.945\textwidth]{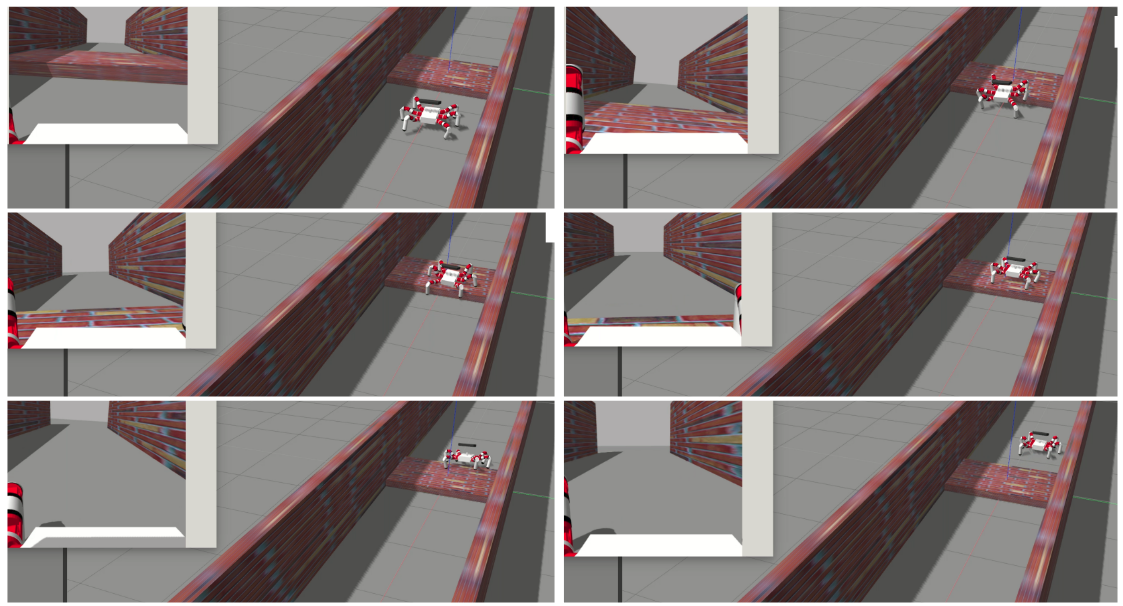}
	\caption{Depiction of the trained meta-policy running on the rough terrain environment. The legged robot learns to sequence two behaviours to traverse the obstacle present in the middle of the environment. The top left corner of each frame shows the camera view of the robot. This camera images are being used by the robot to decide which gait to use at a specific time step. }	
	\label{fig:snakemonster_episode}
\end{figure*}
\subsection{Meta-level control policy}
Our meta-level policy learns an appropriate sequence of the low-level control policies, based on camera observation of the surrounding environment. The action space of the meta-policy corresponds to which low-level policy to execute i.e., $\pi_1 $or $\pi_2$ discussed in Section~\ref{section:lowlevelpolicies}. 
The meta-level policy thus selects a low-level control policy to execute, and the robot executes this policy for $N$ time steps. The meta-policy then subsequently chooses which policy to run, as depicted in Figure~\ref{fig:arch}.

Given the high dimensional observation space of the robots' states considered in this paper (raw camera pixels), we utilize a DQN meta-policy that uses a neural network as a function approximator for the Q-values. The network is basically composed of  $3$ convolutional layers, followed by a $2$ fully connected layers. The the outputs of this network are the meta-level Q-values of running the low-level policies forward for $N$ time steps. 
Note that we use the same architecture of the meta policy for both of the robots we consider in this paper.


%

\subsection{Implementation Details}
Below we provide details of the training setup. The training settings are identical for both low-level control policies and the meta-policy. 

\subsubsection{Reward Function}
The robot's objective is to navigate the largest distance possible without running into the walls, given a particular starting point. This can be framed as a reward maximization problem typical to reinforcement learning. This translates to a positive reward signal proportional to the distance moved by the robot's centre of mass (+5 for moving forward, and +1 for turning left or right), with a highly negative penalty on collisions with obstacles (-200). We use the ground truth position of the robot (relative to its starting point) to determine the distance traveled by the robot, and provide a reward signal directly proportional to this distance. We detect collisions with the obstacles using the distance to obstacle measured by a LIDAR mounted on the robots (only used during training), and penalize such collisions with a reward value of $-200$. 

We note that the reward functions for the low-level policy and the meta-policy are different. The low level policy collects an immediate reward after executing each step, whereas the meta-policy uses the cumulative reward collected over N steps of execution of the low-level policy. Note also that we do not require additional reward functions (such as intrinsic motivation) in contrast with \cite{kulkarni2016hierarchical}.

\subsubsection{Training Details}

During training, we utilize a number of techniques that have proved useful in training reinforcement learning agents. Specifically, we use experience replay \cite{Lin:1992:SRA:139611.139620} of memory size of $1000000$, with a burn-in of $50000$ time steps. We also maintain a target Q-network for training stability as in double Deep Q-Networks \cite{DBLP:journals/corr/HasseltGS15}. The target network is updated every $10000$ time steps. We train our DQN using the Adam optimizer with a learning rate of $0.00025$. We use a discount factor of $0.99$, and gradient clipping of magnitude $1$. Futhermore, we make use of a decaying epsilon greedy policy during training, to ensure sufficient exploration of states and actions.

\section{Results}
\label{results}

In this section, we describe the results of our framework applied to two robots; a legged robot and a mobile robot. We show results of the meta-policy sequencing a set of low-level policies for each of the robots. We demonstrate the ability of meta-level control policies to navigate the robots through the respective environments based only on monocular camera feedback.


\subsection{Legged-Robot Navigation}
The legged robot that has to switch between gaits to traverse across a rough terrain. The low-level policies in this scenario or two CPG gaits. The first gait $\pi_1$ is characterized by high steps and low forward speed. The second gait $\pi_2$ is characterized by low steps and high forward speed. We visualize one such episode in Fig.~\ref{fig:snakemonster_episode}, where we depict the progress of the robot at $6$ varying time steps. Note the robots ability to sequence both gaits to traverse the high obstacle in the middle of path. For a video of the legged robot navigating performing this task, check the \href{https://drive.google.com/file/d/1uGG9WQInaeF39HZVEUZs8u_vfb-5Iylc/view?usp=sharing}{following link}.

\begin{figure}[t]
	\includegraphics[width=.5\textwidth]{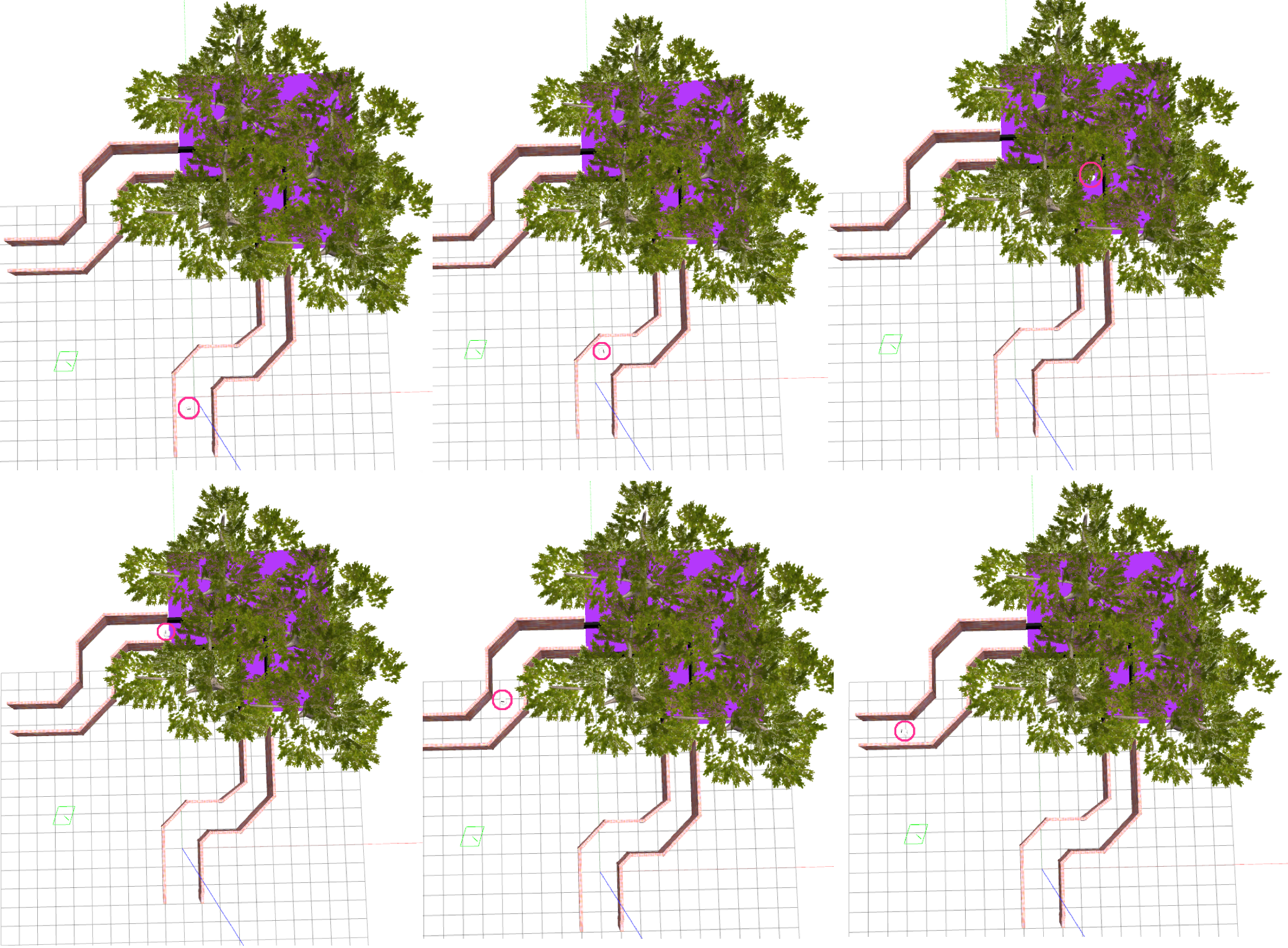}
	\caption{Depiction of the trained meta-policy running on the compound environment. The robot is circled by a pink circle for visibility. Notice the robot able to traverse a relatively long complex path with several turns. }	
	\label{fig:pinky}
\end{figure}

\subsection{Wheeled Robot Navigation}
The wheeled mobile robot (turtlebot) has to navigate in maze-like environment without colliding with obstacles. 
The turtlebot is trained separately on two elementary environments: environment $1$ shown in Fig.~\ref{fig:envs_turtlebot}-left and environment $2$ shown Fig.~\ref{fig:envs_turtlebot}-center. The learned policies in these environments are $\pi_1$ and $\pi_2$ respectively.  The results for the turtle-bot navigating the environment 2 may be viewed at \href{https://drive.google.com/open?id=1s1o0Rh17yaG2uYGGmOuCnE48fWc2v_Oi}{this link}.


\begin{figure}[t]
	\includegraphics[width=.5\textwidth]{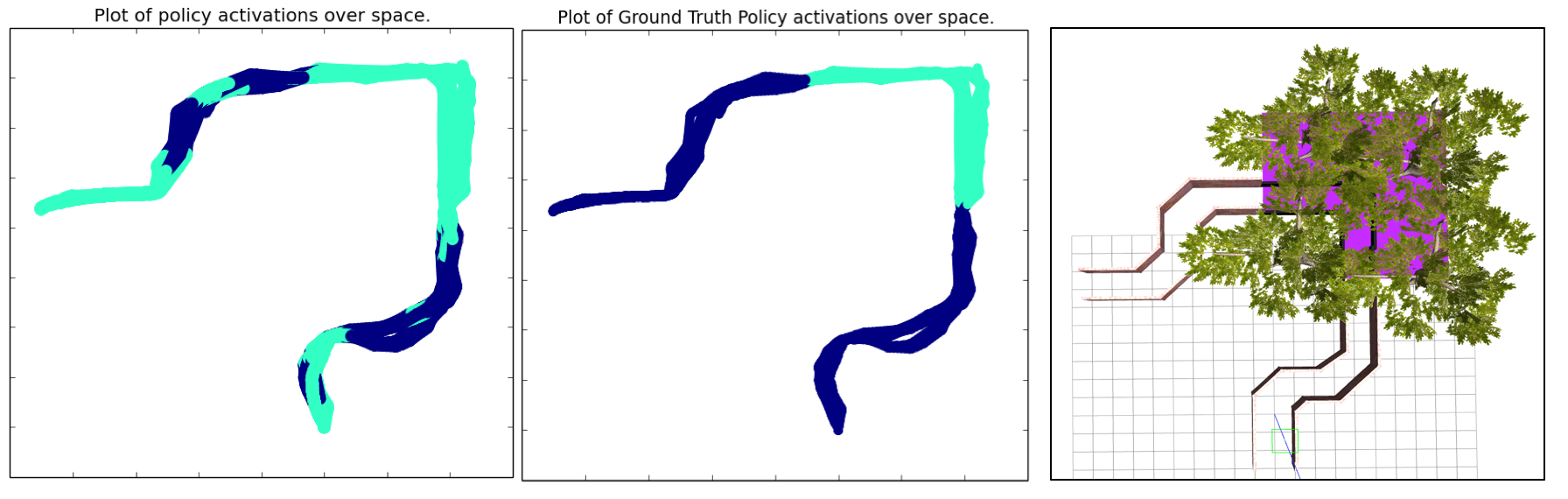}
	\caption{The human constructed deterministic policy (center) is to use $\pi_1$ (depicted in dark blue) in areas with bricks, and $\pi_2$ (depicted in light green) in areas with black walls / purple floors. The learned meta-policy chooses the assignment of policies depicted on the left. While there are areas in which this learned meta-policy differs from the human deterministic policy, the meta-policy is able to do better by switching between different policies.}
	\label{fig:policyact}
\end{figure}

We then learn a new policy $\Omega$ that sequences the policies $\pi_1$ and $\pi_2$ by training on a compound environment that is shown in Fig.~\ref{fig:envs_turtlebot}-right. Note that here, $\pi_1$ performs poorly on environment $2$, and $\pi_2$ performs poorly on environment $1$. This disparity in performances is because the low-level policies are not trained on the opposite environments. The meta-policy learns and exploits this performance in different environments of respective low-level policies. 
\subsection{Discussions}
Intuitively, the meta-policy should learn to apply $\pi_1$ in the area of the compound environments with bricks, and apply $\pi_2$ in the area of the environment with black walls and purple flooring. Pictorially, we can represent this in Fig.~\ref{fig:policyact} center.  Upon training the meta policy to sequence the policies $\pi_1$ and $\pi_2$, we plot the  low-level policies chosen by the meta-policy as a function of space, as shown in Fig.~\ref{fig:policyact}. These \textit{policy activations} provide us insight into how the meta-policy functions. 
As is visible from Fig.~\ref{fig:policyact}-left, we see that the learned meta-policy learns to choose policy $\pi_2$ in environment $2$, consistent with the performance of $\pi_2$ in its training environment.

\begin{table}[]
\centering
\caption{Average reward and task success rate of the our meta-policy versus hand-crafted and learnt baselines}
\label{my-label}
    \resizebox{0.48\textwidth}{!}{
	\begin{tabular}{@{}ccc@{}}
\toprule
Algorithm & \begin{tabular}[c]{@{}c@{}}Average Cumulative Reward\\ (over 10 episodes)\end{tabular} & \begin{tabular}[c]{@{}c@{}}Percentage of\\ Successful Trials\end{tabular} \\ \midrule
\textbf{Meta-level Policy (ours)} & \textbf{988.5} & \textbf{80\%} \\ \midrule
\begin{tabular}[c]{@{}c@{}}Hand-crafted \\ meta-level Policy\end{tabular} & 669.3 & 40\% \\ \midrule
\begin{tabular}[c]{@{}c@{}}Low-level Policy \\ (Compound Env.)\end{tabular} & 408.9 & 0\% \\ \midrule
\begin{tabular}[c]{@{}c@{}}Low-level Policy\\ (Elementary Env. 1)\end{tabular} & 317.9 & 0\% \\ \midrule
\begin{tabular}[c]{@{}c@{}}Low-level Policy\\ (Elementary Env. 2)\end{tabular} & - & - \\ \bottomrule
\end{tabular}}
\end{table}

Further, we see that outside of the black walled area with purple flooring, there are occasions when the meta-policy chooses to alternate between $\pi_1$ and $\pi_2$. Observing the video of the turtlebot in Gazebo while it is executing the behavior, we realize the following interesting behavior of the meta-policy. 

The low-level policy $\pi_2$ traverses environment~1, by slowly switching between left and right actions, without moving \textit{forward}. However, it is able to pass environment~1. Since the forward action has a larger velocity provided to it, the meta-policy learns that such slow, careful behavior is better for portions of environment~1 with turns in it. 
In regions of the environment~1 that are straight, the meta-policy learns to exploit $\pi_1$, which tends to go straight when possible. This modulation of which policy is active as chosen by the meta policy, is visible in Fig.~\ref{fig:policyact}-left. 

This interesting behavior guarantees that the meta-policy is able to traverse the entire path, while neither of the low-level policies can traverse the entire path. Further, the low-level policies are not perfect, and tend to crash occasionally in their own environments. This implies that the human deterministic policy, that simply calls each policy in its respective environment, is not guaranteed to traverse the entire path.
The meta-policy, however, smartly interleaves the two low level policies to traverse the entire path with reasonably good probability, as depicted in Figure \ref{fig:pinky}.

We report the average reward obtained over $10$ episodes in Table~\ref{my-label}. The human deterministic policy is able to achieve an average cumulative reward of $669.3$, while the learned meta-policy achieves an average cumulative reward of $988.5$. This significant difference in the rewards is due to the ability of the meta policy to switch between low-level control policies to nearly guarantee the robot traverses the entire path.

We show that a single low level policy is unable to traverse the entire path, thus necessitating the use of a meta-policy in the  \href{https://drive.google.com/open?id=1T0JDRXRYTs1sZJIgfnajQG6OQqiAQNkg}{following link}. 
Next, we show the results of the handcrafted deterministic policy, which sometimes fails to traverse the path, in the following link \href{https://drive.google.com/open?id=1eiYpuW45ApuSWHDpji0-AJsQqk89OwrT}{following link}
Finally, we show the results of the learned meta-policy, which manages to traverse the path and achieve higher rewards by selecting alternating control policies to use, in the  \href{https://drive.google.com/open?id=1WGr4LFwP54rDVwQQoKO-kosdKO7lhIoI}{following link}. 
Please note the video has annotations of which low-level control policy is being invoked, in the background on the left. Meta-action $0$ refers to $\pi_1$, and meta-action $1$ refers to $\pi_2$. 

\begin{figure}[t]
    \centering
	\includegraphics[width=.40\textwidth]{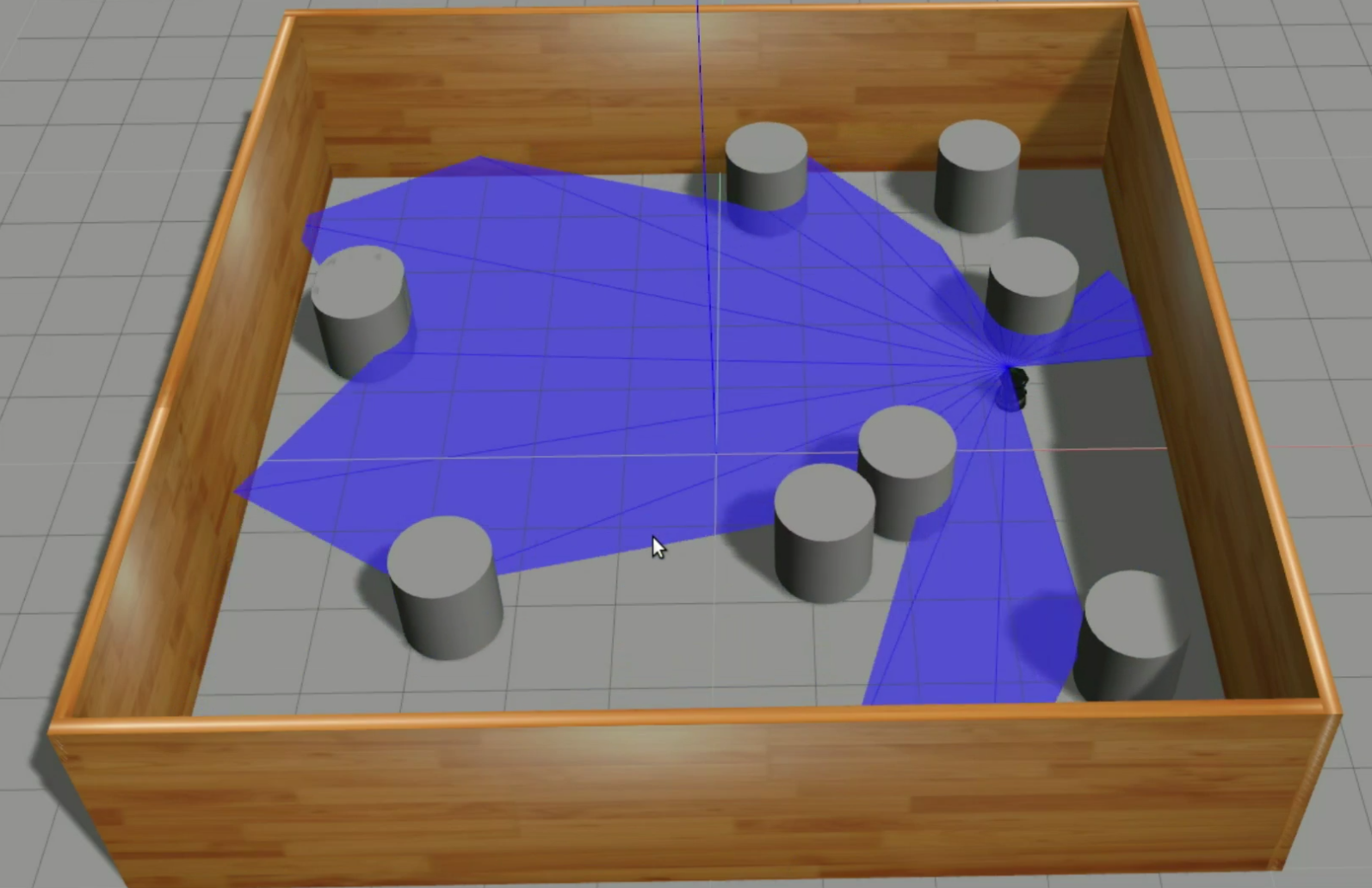}
	\caption{Depiction of the task of navigating a wheeled robot in a dynamic environment using our framework.}
	\label{fig:dyenv}
\end{figure}
\subsection{Wheeled-Robot Navigation in Dynamic Environments}
To explore the capability of our framework to accommodate sensor modalities other than cameras such as LIDARs, and its applicability in dynamic environments, we test our framework on the task of navigating a wheeled mobile robot in an environment with combined static and dynamic obstacles. We depict the robot and the dynamic environment used in this task in Fig.\ref{fig:dyenv}. The results of the successful navigation of the robot can be seen in the video in the 
\href{https://drive.google.com/open?id=1bhIJEapQVMfoQtu1npYMJRyFnrivlxji}{following link}. 
\section{Conclusion and Future Work}
\label{conclusion}

In this paper, we present an approach to both learn and sequence robot behaviors, applied to the problem of visual navigation of mobile robots. 
The layered representation of control policies that we employ allows the robot to adapt to changes in the environment, and select the low-level behavior most appropriate for the current situation, enabling significantly improved task performance. The meta-level policies that we learn are agnostic to the nature of the low-level behaviors used, enabling the use of both hand-crafted as well as learnt policies. In addition, these meta-level policies may also be trained with other input modalities as well. These features exemplify why maintaining a hierarchy of control policies is a potent tool in enabling robots with more autonomous capabilities.

For future work, we would like to explore how this farmework may be modified to deal with learning both levels of policies jointly. This would allow adapting the low-level behaviors online, enabling further capabilities of robots.

\bibliographystyle{IEEEtran}
\bibliography{references}

\begin{thebibliography}{10}
\providecommand{\url}[1]{#1}
\csname url@rmstyle\endcsname
\providecommand{\newblock}{\relax}
\providecommand{\bibinfo}[2]{#2}
\providecommand\BIBentrySTDinterwordspacing{\spaceskip=0pt\relax}
\providecommand\BIBentryALTinterwordstretchfactor{4}
\providecommand\BIBentryALTinterwordspacing{\spaceskip=\fontdimen2\font plus
\BIBentryALTinterwordstretchfactor\fontdimen3\font minus
  \fontdimen4\font\relax}
\providecommand\BIBforeignlanguage[2]{{%
\expandafter\ifx\csname l@#1\endcsname\relax
\typeout{** WARNING: IEEEtran.bst: No hyphenation pattern has been}%
\typeout{** loaded for the language `#1'. Using the pattern for}%
\typeout{** the default language instead.}%
\else
\language=\csname l@#1\endcsname
\fi
#2}}

\bibitem{Nicolescu:2002:HAB:544741.544798}
\BIBentryALTinterwordspacing
M.~N. Nicolescu and M.~J. Matari\'{c}, ``A hierarchical architecture for
  behavior-based robots,'' in \emph{Proceedings of the First International
  Joint Conference on Autonomous Agents and Multiagent Systems: Part 1}, ser.
  AAMAS '02.\hskip 1em plus 0.5em minus 0.4em\relax New York, NY, USA: ACM,
  2002, pp. 227--233. [Online]. Available:
  \url{http://doi.acm.org/10.1145/544741.544798}
\BIBentrySTDinterwordspacing

\bibitem{richter2012robotic}
M.~Richter, Y.~Sandamirskaya, and G.~Sch{\"o}ner, ``A robotic architecture for
  action selection and behavioral organization inspired by human cognition,''
  in \emph{Intelligent Robots and Systems (IROS), 2012 IEEE/RSJ International
  Conference on}.\hskip 1em plus 0.5em minus 0.4em\relax IEEE, 2012, pp.
  2457--2464.

\bibitem{10.3389/fncom.2014.00062}
\BIBentryALTinterwordspacing
G.~Neumann, C.~Daniel, A.~Paraschos, A.~Kupcsik, and J.~Peters, ``Learning
  modular policies for robotics,'' \emph{Frontiers in Computational
  Neuroscience}, vol.~8, p.~62, 2014. [Online]. Available:
  \url{http://journal.frontiersin.org/article/10.3389/fncom.2014.00062}
\BIBentrySTDinterwordspacing

\bibitem{mnih2015human}
V.~Mnih, K.~Kavukcuoglu, D.~Silver, A.~A. Rusu, J.~Veness, M.~G. Bellemare,
  A.~Graves, M.~Riedmiller, A.~K. Fidjeland, G.~Ostrovski, \emph{et~al.},
  ``Human-level control through deep reinforcement learning,'' \emph{Nature},
  vol. 518, no. 7540, pp. 529--533, 2015.

\bibitem{mnih2016asynchronous}
V.~Mnih, A.~P. Badia, M.~Mirza, A.~Graves, T.~Lillicrap, T.~Harley, D.~Silver,
  and K.~Kavukcuoglu, ``Asynchronous methods for deep reinforcement learning,''
  in \emph{International Conference on Machine Learning}, 2016, pp. 1928--1937.

\bibitem{silver2016mastering}
D.~Silver, A.~Huang, C.~J. Maddison, A.~Guez, L.~Sifre, G.~Van Den~Driessche,
  J.~Schrittwieser, I.~Antonoglou, V.~Panneershelvam, M.~Lanctot,
  \emph{et~al.}, ``Mastering the game of go with deep neural networks and tree
  search,'' \emph{Nature}, vol. 529, no. 7587, pp. 484--489, 2016.

\bibitem{koutnik2014evolving}
J.~Koutn{\'\i}k, J.~Schmidhuber, and F.~Gomez, ``Evolving deep unsupervised
  convolutional networks for vision-based reinforcement learning,'' in
  \emph{Proceedings of the 2014 Annual Conference on Genetic and Evolutionary
  Computation}.\hskip 1em plus 0.5em minus 0.4em\relax ACM, 2014, pp. 541--548.

\bibitem{Konidaris11d}
\BIBentryALTinterwordspacing
G.~Konidaris, S.~Kuindersma, R.~Grupen, and A.~Barto, ``Cst: Constructing skill
  trees by demonstration,'' in \emph{Proceedings of the ICML Workshop on New
  Developments in Imitation Learning}, July 2011. [Online]. Available:
  \url{http://lis.csail.mit.edu/pubs/konidaris-icmlws11.pdf}
\BIBentrySTDinterwordspacing

\bibitem{Sutton99betweenmdps}
R.~Sutton, D.~Precup, and S.~Singh, ``Between mdps and semi-mdps: A framework
  for temporal abstraction in reinforcement learning,'' \emph{Artificial
  Intelligence}, vol. 112, pp. 181--211, 1999.

\bibitem{kulkarni2016hierarchical}
T.~D. Kulkarni, K.~Narasimhan, A.~Saeedi, and J.~Tenenbaum, ``Hierarchical deep
  reinforcement learning: Integrating temporal abstraction and intrinsic
  motivation,'' in \emph{Advances in Neural Information Processing Systems},
  2016, pp. 3675--3683.

\bibitem{DBLP:journals/corr/AndreasKL16}
\BIBentryALTinterwordspacing
J.~Andreas, D.~Klein, and S.~Levine, ``Modular multitask reinforcement learning
  with policy sketches,'' \emph{CoRR}, vol. abs/1611.01796, 2016. [Online].
  Available: \url{http://arxiv.org/abs/1611.01796}
\BIBentrySTDinterwordspacing

\bibitem{tlpkICRA17}
\BIBentryALTinterwordspacing
T.~L.-P. Leslie Pack~Kaelbling, ``Learning composable models of parameterized
  skills,'' in \emph{IEEE Conference on Robotics and Automation (ICRA)}, 2017.
  [Online]. Available: \url{http://lis.csail.mit.edu/pubs/lpk/ICRA17.pdf}
\BIBentrySTDinterwordspacing

\bibitem{xie2017towards}
L.~Xie, S.~Wang, A.~Markham, and N.~Trigoni, ``Towards monocular vision based
  obstacle avoidance through deep reinforcement learning,'' \emph{arXiv
  preprint arXiv:1706.09829}, 2017.

\bibitem{2017arXiv171009767F}
K.~{Frans}, J.~{Ho}, X.~{Chen}, P.~{Abbeel}, and J.~{Schulman}, ``{Meta
  Learning Shared Hierarchies},'' \emph{ArXiv e-prints}, Oct. 2017.

\bibitem{DBLP:journals/corr/WangFL15}
\BIBentryALTinterwordspacing
Z.~Wang, N.~de~Freitas, and M.~Lanctot, ``Dueling network architectures for
  deep reinforcement learning,'' \emph{CoRR}, vol. abs/1511.06581, 2015.
  [Online]. Available: \url{http://arxiv.org/abs/1511.06581}
\BIBentrySTDinterwordspacing

\bibitem{sartoretticentral}
G.~Sartoretti, S.~Shaw, K.~Lam, N.~Fan, M.~Travers, and H.~Choset, ``Central
  pattern generator with inertial feedback for stable locomotion and climbing
  in unstructured terrain.''

\bibitem{zamora2016extending}
I.~Zamora, N.~G. Lopez, V.~M. Vilches, and A.~H. Cordero, ``Extending the
  openai gym for robotics: a toolkit for reinforcement learning using ros and
  gazebo,'' \emph{arXiv preprint arXiv:1608.05742}, 2016.

\bibitem{1606.01540}
G.~Brockman, V.~Cheung, L.~Pettersson, J.~Schneider, J.~Schulman, J.~Tang, and
  W.~Zaremba, ``Openai gym,'' 2016.

\bibitem{Quigley09}
M.~Quigley, B.~Gerkey, K.~Conley, J.~Faust, T.~Foote, J.~Leibs, E.~Berger,
  R.~Wheeler, and A.~Ng, ``Ros: an open-source robot operating system,'' in
  \emph{Proc. of the IEEE Intl. Conf. on Robotics and Automation (ICRA)
  Workshop on Open Source Robotics}, Kobe, Japan, May 2009.

\bibitem{koenig2006gazebo}
N.~Koenig and A.~Howard, ``Gazebo-3d multiple robot simulator with dynamics,''
  2006.

\bibitem{Lin:1992:SRA:139611.139620}
\BIBentryALTinterwordspacing
L.-J. Lin, ``Self-improving reactive agents based on reinforcement learning,
  planning and teaching,'' \emph{Mach. Learn.}, vol.~8, no. 3-4, pp. 293--321,
  May 1992. [Online]. Available: \url{https://doi.org/10.1007/BF00992699}
\BIBentrySTDinterwordspacing

\bibitem{DBLP:journals/corr/HasseltGS15}
\BIBentryALTinterwordspacing
H.~van Hasselt, A.~Guez, and D.~Silver, ``Deep reinforcement learning with
  double q-learning,'' \emph{CoRR}, vol. abs/1509.06461, 2015. [Online].
  Available: \url{http://arxiv.org/abs/1509.06461}
\BIBentrySTDinterwordspacing

\end{thebibliography}

\end{document}